\def\year{2020}\relax
\documentclass[letterpaper]{article} % DO NOT CHANGE THIS
\usepackage{aaai20}  % DO NOT CHANGE THIS
\usepackage{times}  % DO NOT CHANGE THIS
\usepackage{helvet} % DO NOT CHANGE THIS
\usepackage{courier}  % DO NOT CHANGE THIS
\usepackage[hyphens]{url}  % DO NOT CHANGE THIS
\usepackage{graphicx} % DO NOT CHANGE THIS
\usepackage{graphicx}  % DO NOT CHANGE THIS
\usepackage{amsmath}
\usepackage{amsthm}
\usepackage{amssymb}
\usepackage{multirow}
\title{On the Learning Property of Logistic and Softmax Losses for Deep Neural Networks}
\author{Xiangrui Li, Xin Li, Deng Pan, Dongxiao Zhu\thanks{Corresponding author}\\ % All authors must be in the same font size and format. Use \Large and \textbf to achieve this result when breaking a line
Department of Computer Science\\
Wayne State University\\ %If you have multiple authors and multiple affiliations
\{xiangruili, xinlee, pan.deng, dzhu\}@wayne.edu
}

\begin{document}

\maketitle

\begin{abstract}
Deep convolutional neural networks (CNNs) trained with logistic and softmax losses have made significant advancement in visual recognition tasks in computer vision. When training data exhibit class imbalances, the class-wise reweighted version of logistic and softmax losses are often used to boost performance of the unweighted version. In this paper, motivated to explain the reweighting mechanism, we explicate the learning property of those two loss functions by analyzing the necessary condition (e.g., gradient equals to zero) after training CNNs to converge to a local minimum. The analysis immediately provides us explanations for understanding (1) quantitative effects of the class-wise reweighting mechanism: deterministic effectiveness for binary classification using logistic loss yet indeterministic for multi-class classification using softmax loss; (2) disadvantage of logistic loss for single-label multi-class classification via one-\textit{vs.}-all approach, which is due to the averaging effect on predicted probabilities for the negative class (e.g., non-target classes) in the learning process. With the disadvantage and advantage of logistic loss disentangled, we thereafter propose a novel reweighted logistic loss for multi-class classification. Our simple yet effective formulation improves ordinary logistic loss by focusing on learning hard non-target classes (target \textit{vs.} non-target class in one-\textit{vs.}-all) and turned out to be competitive with softmax loss. We evaluate our method on several benchmark datasets to demonstrate its effectiveness.
\end{abstract}

\section{Introduction}
Deep convolutional neural networks (CNNs) trained with logistic or softmax losses (LGL and SML respectively for brevity), e.g., logistic or softmax layer followed by cross-entropy loss, have achieved remarkable success in various visual recognition tasks  \cite{lecun2015deep,alexnet,resnet,vgg,inception}. The success mainly accredits to CNN's merit of high-level feature learning and loss function's differentiability and simplicity for optimization. When training data exhibit class imbalances, training CNNs with gradient descent is biased towards learning majority classes in the conventional (unweighted) loss, resulting in performance degradation for minority classes. To remedy this issue, the class-wise reweighted loss is often used to emphasize the minority classes that can boost the predictive performance without introducing much additional difficulty in model training \cite{cui2019class,huang2016learning,mahajan2018exploring,wang2017learning}. A typical choice of weights for each class is the inverse-class frequency.

A natural question then to ask is \textit{what roles are those class-wise weights playing in CNN training using LGL or SML that lead to performance gain?} Intuitively, those weights make tradeoffs on the predictive performance among different classes. In this paper, we answer this question quantitatively in a set of equations that tradeoffs are on the model predicted probabilities produced by the CNN models. Surprisingly, effectiveness of the reweighting mechanism for LGL is rather different from SML. Here, we view the conventional (e.g., no reweighting) LGL or SML as a special case where all classes are weighted equally.

As these tradeoffs are related to the logistic and softmax losses, answering the above question actually leads us to answering a more fundamental question about their learning behavior: \textit{what is the property that the decision boundary must satisfy when models are trained?} To our best knowledge, this question has not been investigated systematically, despite logistic and softmax losses are extensively exploited in deep leaning community.

While SML can be viewed as a multi-class extension of LGL for binary classification, LGL is a different learning objective when used in multi-class classification \cite{bishop2006pattern}. From the perspective of learning structure of data manifold as pointed out in \cite{belkin2006manifold,bishop2006pattern,dong2019single}, SML treats all class labels equally and poses a competition between true and other class labels for each training sample, which may distort data manifold; for LGL, the one-\textit{vs.}-all approach it takes avoids this limitation as it models each target class independently, which may better capture the in-class structure of data. Though LGL enjoys such merits, it is rarely adopted in existing CNN models. The property that LGL and SML decision boundaries must satisfy further reveals the difference between LGL and SML (see Eq. (\ref{eq:keyEq2}), (\ref{eq:binaykeyEq}) with analysis). If used for the multi-class classification problem, we can identify two issues for LGL. Compared with SML, LGL may introduce data imbalance, which can degrade model performance as sample size plays an important role in determining decision boundaries. More importantly, since the one-\textit{vs.}-all approach in LGL treats all other classes as the negative class, which is of a multi-modal distribution \cite{li2018robust,li2018multinomial}, the averaging effect of the predicted probabilities of LGL can hinder learning discriminative feature representations to other classes that share some similarities with the target class.

Our contribution can be summarized as follows:
\begin{itemize}
	\item We provide a theoretical derivation on the relation among sample’s predicted probability (once CNN is trained), class weights in the loss function and sample size in a system of equations. Those equations explaining the reweighting mechanism are different in effect for LGL and SML. 
	\item We depict the learning property for LGL and SML for classification problems based on those probability equations. Under mild conditions, the expectation of model predicted probabilities must maintain a relation specified in Eq (\ref{eq:keyEq2}).
	\item We identify that the multi-modality neglect problem in LGL is the main obstacle for LGL in multi-class classification. To remedy this problem, we propose a novel learning objective, in-negative class reweighted LGL, as a competitive alternative for LGL and SML.
	\item We conduct experiments on several benchmark datasets to demonstrate the effectiveness of our method. 
\end{itemize}

\section{Related Work}
With recent explosion in computational power and availability of large scale image datasets, deep learning models have repeatedly made breakthroughs in a wide spectrum of tasks in computer vision \cite{lecun2015deep,goodfellow2016deep}. Those advancements include new CNN architectures for image classification\cite{alexnet,resnet,vgg,inception}, objective detection and segmentation \cite{ren2015faster,ronneberger2015u}, new loss functions \cite{dong2019single,zhang2018generalized} and effective training techniques to improve CNN performance \cite{dropout,ioffe2015batch}. 

In those supervised learning problems, CNNs are mostly trained with loss functions such as LGL and SML. In practice, class imbalance naturally emerges in real-world data and training CNN models directly on those datasets may lead to poor performance. This phenomenon is referred as the imbalanced learning problem \cite{he2008learning}. To tackle this problem, cost-sensitive method \cite{elkan2001cs,zhou2010multi} is the widely-adopted approach in current training practices as they don't introduce any obstacles in the backpropagation algorithm. One of the most popular methods is class-wise reweighting loss function based on LGL and SML. For example, \cite{huang2016learning,wang2017learning} reweight each class by its inverse-class frequency. In some long-tailed datasets, a smoothed version of weights is adopted \cite{mahajan2018exploring,mikolov2013bow}, which emphasizes less on minority classes, such as the square root of inverse-class frequency. More recently, \cite{cui2019class} proposed a weighting strategy based on the calculation of effective sample size. In the context of learning from noisy data, \cite{zhang2018generalized} provides analysis on the weighted SGL showing close connection to the mean absolute error (MAE) loss. However, what role class-wise weights play in LGL and SML is not explained in previous works. In this paper, we provide a theoretical explication on how the weights control the tradeoffs among model predictions.

If we decompose the multi-class classification as multiple binary classification sub-tasks, LGL can also be used as the objective function via one-\textit{vs.}-all approach \cite{hastie2005elements,bishop2006pattern}, which is however rarely adopted in existing works of deep learning. Motivated to understand class-wise reweighted LGL and SML, our analysis further leads us to a more profound discovery in the properties of decision boundaries for LGL and SML. Previous work in \cite{dong2019single} showed that the learning objective using LGL is quite different from SML as each class is learned independently. They identified the negative class distraction (NCD) phenomenon that might be detrimental to model performance when using LGL in multi-class classification. From our analysis, the NCD problem can be partially explained that LGL treats the negative class (e.g., non-target classes) as a single class and ignores its multi-modality. If there exists one non-target class that share some similarity with the target class, CNN trained with LGL may make less confident predictions for that non-target class (e.g., probability of belonging to the negative class is small) as its predicted probabilities are averaged out due to other non-target classes with confident predictions. Consequently, samples from that specific non-target class can be misclassified into the target class, resulting in large predictive error.

\section{Analysis on LGL and SML} 
In this section, we provide a theoretical explanation for the class-wise weighting mechanism and depict the learning property of LGL and SML losses.

\noindent\textbf{Notation} Let $D = \{(\boldsymbol{x}_i, y_i)\}_{i=1}^{N}$ be the set of training samples of size $N$, where $\boldsymbol{x}_i \in \textbf{R}^p$ is the $p$-dimensional feature vector and $y_i = k (k=0,\cdots, K-1)$ is the true class label, and $S_k= \{(\boldsymbol{x}_i, y_i): y_i=k\}$ the subset of $D$ for the $k$-th class. The bold $\boldsymbol{y}_i=(y_i^0,\cdots,y_i^{K-1})$ is used to represent the one-hot encoding for $y_i$: $y_i^k=1$ if $y_i = k$, $0$ otherwise. $N_k = |S_k|(k=0,\cdots,K-1)$ is used to represent sample size for the $k$-th class and hence $\sum_k N_k=N$. The maximum size is denoted as $N_{\max}= \max_{k=0,\cdots,K-1}(N_k)$.

\subsection{Preliminaries}
For classification problem, the probability for a sample $\boldsymbol{x}$ belonging to one class is modeled by logistic (e.g., sigmoid) for binary classification 
\[p(y=1|\boldsymbol{x};\boldsymbol{\theta}) = \frac{1}{1+\exp(-z)},\]
\[ p(y=0|\boldsymbol{x};\boldsymbol{\theta}) =1-p(y=1|\boldsymbol{x}),\]
and by softmax for multi-class classification
\[p(y=k|\boldsymbol{x};\boldsymbol{\theta}) = \frac{\exp(z_k)}{\sum_{j=0}^{K-1}\exp(z_j)},\]
where all $z$'s are the logits for $\boldsymbol{x}$ modeled by CNN with parameter vector $\boldsymbol{\theta}$. It is worth noting that softmax is equivalent to logistic in binary classification as can be seen from
\[p(y=1|\boldsymbol{x}) = \frac{\exp(z_1)}{\exp(z_0)+\exp(z_1)} = \frac{1}{1+\exp(-(z_1-z_0))}.\]

Hence, without loss of generality, we write class-wise reweighted LGL ($K=2$) and SML ($K\geq 3$) in a unified form as follows
\begin{equation} \label{eq:RSML}
L(\boldsymbol{\theta}) = -\sum_{k=0}^{K-1}\lambda_k \sum_{i_k\in S_k}  \log f_k(\boldsymbol{\theta};\boldsymbol{x}_{i_k}) 
= -\sum_{k=0}^{K-1}\lambda_k L_k(\boldsymbol{\theta}),
\end{equation}
where each $f_k(\boldsymbol{\theta};\boldsymbol{x}_i) = p(y_i =k|\boldsymbol{x}_i)$ is the CNN predicted probability of sample $\boldsymbol{x}_i$ belonging to the $k$-th class; $\lambda$s are weight parameters to control each class's contribution in the loss. When all $\lambda$s are equal, $L(\boldsymbol{\theta})$ is the conventional cross-entropy loss and minimizing it is equivalent to maximizing likelihood. If the training data are imbalanced, a different setup of $\lambda$s is used, usually classes with smaller sizes are assigned with higher weights. Generally, $\lambda$s are treated as hyperparameters and selected by cross-validation.

We emphasize here that using logistic function for multi-class ($K \geq 3$) is a different learning objective from softmax in this case as the classification problem is essentially reformulated as $K$ binary classification sub-problems. 

\subsection{Key Equations for Weights $\lambda$s} 
Assume that CNN's output layer, after convolutional layers, is a fully connected layer of $K$ neurons with bias terms, then the predicted probability for sample $\boldsymbol{x}$ is given by the softmax activation:
\begin{equation} \label{eq:prob}
f_k(\boldsymbol{x}) = \frac{\exp(\boldsymbol{W_k}\boldsymbol{h}_{\boldsymbol{x}}+b_k)}{\sum_{j=1}^{K}\exp(\boldsymbol{W}_j \boldsymbol{h}_{\boldsymbol{x}}+b_j)} \hspace{3mm}(k=0,\cdots, K-1),
\end{equation}
where $\boldsymbol{h}_{\boldsymbol{x}}$ is the feature representation of $\boldsymbol{x}$ extracted from convolutional layers, $\boldsymbol{W}_k$ and $b_k$ are parameters of the $k$-th neuron in the output layer. For notational simplicity, we have dropped $\boldsymbol{\theta}$ in $f_k(\boldsymbol{x})$.

After CNN is trained, we assume that the reweighted SML $L(\boldsymbol{\theta})$ is minimized to local optimum $\boldsymbol{\theta}^*$. By optimization theory, a necessary condition is that the gradient of $L(\boldsymbol{\theta})$ is zero at $\boldsymbol{\theta} = \boldsymbol{\theta}^*$\footnote{More strictly, zero is in the subgradient of $L(\boldsymbol{\theta})$ at $\boldsymbol{\theta}^*$. But this doesn't affect the following analysis.}:
\begin{equation}
\frac{\partial L}{\partial \boldsymbol{\theta}}  =\boldsymbol{0}
\iff   \sum_{k=1}^K \lambda_k \frac{\partial L_k}{\partial \boldsymbol{\theta}}= \boldsymbol{0}.
\end{equation}
We specifically consider $L_1(\boldsymbol{\theta})$ for the $1$-st class with respect to one component $\eta$ of $\boldsymbol{\theta}$. Then with chain rule, the necessary condition above gives:
\begin{equation} \label{eq:necessaryCon}
\begin{split}
&\lambda_1 \frac{\partial L_1}{\partial \eta} + \sum_{k=2}^{K}\lambda_k \frac{\partial L_k}{\partial \eta} = 0 \iff \\
&\lambda_1 \sum_{i_1\in S_1} \frac{1}{f_{1,i_1}} \frac{\partial f_{1,i_1}}{\partial \eta} + \sum_{k=2}^{K} \lambda_k \sum_{i_k\in S_k} \frac{1}{f_{k, i_k}} \frac{\partial f_{k, i_k}}{\partial \eta}=0,
\end{split}
\end{equation}
where we use  $f_{j,i_k} = f_j(\boldsymbol{x}_{i_k})$ given by Eq. (\ref{eq:prob}).

Let $\boldsymbol{\sigma}(\boldsymbol{z})$ be the softmax function of $\boldsymbol{z}=(z_1,\cdots,z_K)$ with each component $\boldsymbol{\sigma}(z_k) = \exp(z_k)/\sum_i \exp(z_i)$, its derivative is 
\begin{equation} \label{eq:smDir}
\frac{\partial \boldsymbol{\sigma}(z_k)}{\partial z_i}= \begin{cases}
\boldsymbol{\sigma}(z_k) (1-\boldsymbol{\sigma}(z_k)), &i=k\\
-\boldsymbol{\sigma}(z_k) \boldsymbol{\sigma}(z_i), &i\ne k.
\end{cases}
\end{equation}

Denoting $a_{j,i_k} = \boldsymbol{W_j}\boldsymbol{h}_{\boldsymbol{x}_{i_k}}+b_j $ as the $j$-th logit in Eq. (\ref{eq:prob}) for sample $\boldsymbol{x}_{i_k}$, then $f_{j,i_k} = \boldsymbol{\sigma}(a_{j,i_k}) (j=0,\cdots, K-1)$. Again with chain rule and Eq. (\ref{eq:smDir}):
\begin{equation} \label{eq:chain}
\begin{split}
\frac{\partial f_{k,i_k}}{\partial \eta} &= \sum_{j=1}^{K} \frac{\partial f_{k,i_k}}{\partial a_{j,i_k}} \frac{\partial a_{j,i_k}}{\partial \eta}\\
& = f_{k, i_k}(1-f_{k, i_k}) \frac{\partial a_{k, i_k}}{\partial \eta} - f_{k,i_k} \sum_{j\ne k} f_{j,i_k} \frac{\partial a_{j, i_k}}{\partial \eta}.
\end{split}
\end{equation}

Since Eq. (\ref{eq:necessaryCon}) holds valid for any component $\eta$ of $\boldsymbol{\theta}$, we specifically consider the case when $\eta=b_1$. Therefore we have $\partial a_{1,i_k}/\partial b_1 = 1$ and $ \partial a_{j,i_k}/\partial b_1 = 0 (j=2,\cdots, K)$. Then Eq. (\ref{eq:chain}) becomes:
\begin{equation} \label{eq:bias-dir}
\frac{\partial f_{k,i_k}}{\partial b_1} = \begin{cases} 
f_{1,i_1} (1-f_{1,i_1}) , &k=1\\
-f_{k,i_k} f_{1,i_k}, & k\ne 1.
\end{cases}
\end{equation}

Plug Eq. (\ref{eq:bias-dir}) back into Eq. (\ref{eq:necessaryCon}) and rearrange the terms, we have 
\begin{equation} \label{eq:keyEq}
\lambda_1\sum_{i_1 \in S_1}(1-f_{1,i_1}) = \sum_{k=2}^{K} \lambda_k \sum_{i_k\in S_k} f_{1, i_k}.
\end{equation}

With the same calculations, we can obtain other $K-1$ similar equations, each of which corresponds to one class. Remember $f_{j,i_k}$ is the probability of sample $\boldsymbol{x}_{i_k}$ from the $k$-th class being predicted into the $j$-th class, and Eq. (\ref{eq:keyEq}) reveals the quantitative relation between weights $\lambda$s, model predicted probabilities and training samples. Notice that CNN is often trained with $L_2$ regularization to prevent overfitting. If the bias term $b_k$s are not penalized, Eq. (\ref{eq:keyEq}) still holds valid. Another possible issue is that the calculation relies on the use of bias terms $b_k$ in the output layer. As using bias increases CNN's flexibility and is not harmful to CNN performance, our analysis is still applicable to a wide range of CNN models trained with cross-entropy loss.

We observe in Eq. (\ref{eq:keyEq}), $\sum_{i_1 \in S_1}(1-f_{1,i_1})/N_1$ (approximately) represents the expected probability of CNN incorrectly predicting a sample of class $1$ and $\sum_{i_k\in S_k} f_{1, i_k}/N_k$ the expected probability of CNN misclassifying a sample of class $k (k\ne 1)$ into class $1$. If we assume that the training data can well represent the true data distribution that testing data also follow, the learning property of trained CNN shown in Eq. (\ref{eq:keyEq}) can be generalized to testing data. 

More specifically, since the CNN model is a continuous mapping and the softmax output is bounded between 0 and 1, by the uniform law of large numbers \cite{newey1994large}, we have the following system of $K$ equations \textit{once CNN is trained}: 
\begin{equation} \label{eq:keyEq2}
\begin{cases}
\lambda_0 N_0(1-\bar{p}_{0\rightarrow 0})  \approx \sum_{k\ne 0} \lambda_k N_k \bar{p}_{k \rightarrow 0}\\
\hspace{2cm} \vdots\\
\lambda_{K-1} N_{K-1}(1-\bar{p}_{K-1\rightarrow K-1})  \approx \sum_{k\ne K-1} \lambda_k N_k \bar{p}_{k \rightarrow K-1},
\end{cases}
\end{equation}
where for indices $i$ and $j$, $\bar{p}_{i\rightarrow j}$ represents the expected probability of CNN predicting a sample from class $i$ into class $j$:
\[\bar{p}_{i\rightarrow j} = \text{E}_{\boldsymbol{x} \sim P(\boldsymbol{x}|y=i)} f_j (\boldsymbol{x}), \]
where $P(\boldsymbol{x}|y=i)$ is the true data distribution for the $i$-th class.

\noindent\textbf{Binary Case with LGL} For binary classification problem ($K=2$), Eq. (\ref{eq:keyEq2}) gives us the following relation about CNN predicted probabilities:
\begin{equation}\label{eq:binaykeyEq}
\frac{1-\bar{p}_{0\rightarrow 0}}{\bar{p}_{1\rightarrow 0}} \approx \frac{\lambda_1 N_1}{\lambda_0 N_0}.
\end{equation}

\begin{itemize}
	\item In the conventional LGL where each class is weighted equally ($\lambda_0 = \lambda_1$), Eq. (\ref{eq:binaykeyEq}) becomes $1-\bar{p}_{0\rightarrow 0} = N_1\bar{p}_{1\rightarrow 0}/N_0$. If data exhibit severe imbalance, say $N_0=10 N_1$, then we must have ($\bar{p}_{1\rightarrow 0}<1$) 
	\[\bar{p}_{0\rightarrow 0} = 1- \frac{\bar{p}_{1\rightarrow 0}}{10}>0.9.\]
	If $t = 0.5$ is the decision making threshold, this implies that
	the trained neural network can correctly predict a majority class (e.g., class 0)
	sample, confidently (at least) with probability 0.9, on average. However, for minority class, the predictive performance is more complex which depends on the trained model and data distribution. For example, if two classes can be well separated and the model made very confident predictions, say $\bar{p}_{0\rightarrow 0} =0.98$, then we must have $\bar{p}_{1\rightarrow 1}=0.8$ for the minority class, implying a good predictive performance on class 1. If $\bar{p}_{0\rightarrow 0} =0.92$, then we have $\bar{p}_{1\rightarrow 1}=0.2$. This means the
	predicted probability of a minority sample being minority is 0.2 on average. Hence, the classifier must misclassify most minority samples ($0.2< 0.5$), resulting in very poor predictive accuracy for minority class.
	
	\item If LGL is reweighted using inverse-class frequencies, $\lambda_0 =1/N_0$ and $\lambda_1 = 1/N_1$, the equation above is equivalent to $\bar{p}_{0\rightarrow 0} = 1-\bar{p}_{1\rightarrow 0}=\bar{p}_{1\rightarrow 1}$. Since predictions are made by  $y=\arg\max_i f_i(\boldsymbol{x})$ and $f_1(\boldsymbol{x}) > f_0(\boldsymbol{x})$ means $f_1(\boldsymbol{x})>0.5$, we can have a deterministic relation: if either class 0 or 1 can be well predicted (e.g., $\bar{p}_{i\rightarrow i}>0.5$), reweighting by class inverse frequencies can guarantee performance improvement for the minority class. However, the extent of ``goodness" depends on the separability of the underlying data distributions of the two classes.	
\end{itemize}

\noindent\textbf{Simulations for Eq. (\ref{eq:binaykeyEq})} We conduct simulations under two settings for checking Eq. (\ref{eq:binaykeyEq}). The imbalance ratio is set to 10 in training data ($N_0=1000, N_1=10000$), testing data size is $(1000, 1000)$; both training and testing data follow the same data distribution. As the property only relies on the last fully connected hidden layer, we use the following setup:
\begin{itemize}
	\item Sim1: $P_1(x|y=1) =\mathcal{N}(-1.5, 1)+\mathcal{U}(0,0.5)$, $P_2(x|y=0) = \mathcal{N}(1.5, 1) + \mathcal{U}(-0.5, 0)$. Logistic regression is fitted. $\mathcal{N}$ and $\mathcal{U}$ represents normal and uniform distribution respectively.
	\item Sim2: $P_1(\boldsymbol{x}|y=1) = \mathcal{N}(\boldsymbol{\mu}_1, \boldsymbol{\sigma}_1)$, $P_0(\boldsymbol{x}|y=0) = \mathcal{N}(\boldsymbol{\mu}_0, \boldsymbol{\sigma}_0)$, where $\boldsymbol{\mu}_1 = (0,0,0)$, $\boldsymbol{\mu}_0 = (1,1,1)$, $\boldsymbol{\sigma}_1 = 1.2\boldsymbol{I}$, $\boldsymbol{\sigma}_0 = \boldsymbol{I}$. A one-hidden-layer forward neural network of layer size $(3,10,1)$ with sigmoid activation. 
\end{itemize}
Table \ref{tab:sim} shows simulation results under three $\lambda$ settings. We see from the Table that simulated values match with the theoretical values accurately, demonstrating the correctness of Equation (\ref{eq:binaykeyEq}).
\begin{table}[t]
	\centering
	\begin{tabular}{cccc}  
		\hline
		$\lambda_0$  & $\frac{1}{2}$  & $\frac{N_0}{N_0+N_1}$  & $\frac{2N_0}{(2N_0+N_1)}$ \\
		\hline
		RHS & 10 & 1 & 0.5  \\
		\hline
		\multirow{2}{*}{LHS (Sim1)} & 10.05   & 1.00   & 0.50  \\
		& (1.13) & (0.09) & (0.04)\\
		\multirow{2}{*}{LHS (Sim2)} & 10.12   & 1.01   & 0.50  \\
		& (0.67) & (0.05) & (0.03)\\
		\hline
	\end{tabular}
	\caption{Simulation results (along with standard deviation) for Eq. (\ref{eq:binaykeyEq}) over 100 runs, $\lambda_1 = 1-\lambda_0$. RHS represents theoretical value on the right-hand side of (\ref{eq:binaykeyEq}); LHS the simulated value on the left hand side.}
	\label{tab:sim}
\end{table}

\noindent\textbf{Multi-class Case with SML} Because $\sum_k \bar{p}_{i\rightarrow k} = 1$ and Eq. (\ref{eq:keyEq2}) has $K(K-1)$ variables with only $K$ equations, we can't exactly solve it quantitatively for a relation among those $\bar{p}_{i\rightarrow j}$'s when $K>2$. For the special case when weights are chosen as the inverse-class frequencies $\lambda_k = 1/N_k$, considering for class 1, we have $(1-\bar{p}_{1\rightarrow 1})  \approx \sum_{k\ne 1} \bar{p}_{k \rightarrow 1}$. Multi-class classification $(K>2)$ does not have a deterministic relation as in the binary case, as predictions are made by  $y=\arg\max_i f_i(\boldsymbol{x})$ and we don't have a decisive threshold for decision making (like the 0.5 in binary case). Our findings match the results in \cite{zhou2010multi} in the sense that class-wise reweighting for multi-class is indeterministic. However, our results are solely based on the mathematical property of the backpropagation algorithm from optimization theory whereas \cite{zhou2010multi} is based on decision theory.

\noindent\textbf{Learning property of LGL and SML}
As the class-wise reweighting mechanism is explained in Eq. (\ref{eq:keyEq2}), those equations also reveal the property of decision boundaries for LGL and SML. For comparison, the decision boundary of support vector machine (SVM) \cite{cortes1995svm} is determined by those support vectors that maximize the margin and those samples with larger margin have no effects on the position of decision boundary. On the contrary, all samples have their contribution to the decision boundary in LGL and SML so that their averaged probabilities that the model produces must satisfy  Eq. (\ref{eq:keyEq2}). In particular for the binary case, we can see that if classes are balanced, the model must make correct predictions with equal confidence for the positive and negative classes, on average; whereas for imbalanced data, the decision boundary will be pushed towards the minority class in a position with Eq (\ref{eq:binaykeyEq}) always maintained. Another observation is that if the expectation of model predicted probabilities doesn't match with its mode (e.g skewed distribution), the magnitude of tradeoff between performance of the majority and minority class depends on the direction of skewness. If the distribution of the majority class skews away from the decision boundary, upweighting minority class will boost model performance at a small cost of performance degradation for the majority class than if it skews towards the decision boundary. This implies that estimating the shape of data distribution in the latent feature space and choosing the weights accordingly would be very helpful to improve model overall performance. 

\section{In-negative Class Reweighted LGL}

In this section, we focused on LGL for multi-class classification via one-\textit{vs.}-all approach. In addition to the theoretical merits of LGL mentioned in the introduction section that LGL is capable of better capturing the structure of data manifold than SML, the guarantee of achieving good performance after properly reweighting (e.g., Eg.(\ref{eq:binaykeyEq})) is also desirable as the one-\textit{vs.}-all approach naturally introduces data imbalance issue.

\noindent\textbf{Multi-modality Neglect Problem} In spite of those merits of LGL, it also introduces the multi-modality neglect problem for multi-class classification. Since the expectation of model predicted probability must satisfy Eq (\ref{eq:binaykeyEq}) for LGL, the averaging effect might be harmful for model performance. In the one-\textit{vs.}-all approach, the negative class consists of all the remaining non-target classes, which follows a multi-modal distribution (one modality for each non-target class). LGL treats all non-target classes equally in the learning process. If there is a hard non-target class that shares non-trivial similarity with the target class, its contribution in LGL might be averaged out by other easy non-target classes. In other words, those easy non-target classes (e.g., correctly predicted as the negative class with high probabilities) would compensate the predicted probability of the hard non-target class so that the probabilistic relation in Eq (\ref{eq:binaykeyEq}) is maintained. Consequently, model could incorrectly predict samples from the hard non-target class into the target class, inducing large predictive error for that class. This phenomenon is not desirable as we want LGL to pay more attention on the separation of the target-class with that hard class, meanwhile maintain the separation from the remaining easy non-target classes. 

To this end, we propose an improved version of LGL to reweight each non-target class's contribution within the negative class. Specifically, for the target class $k$ (e.g., positive class, labeled as $y=1$) and all non-target classes (e.g., negative class, labeled as $y=-1$), a two-level reweighting mechanism is applied in LGL, which we term as \textbf{in-negative-class reweighted LGL (LGL-INR)}:
\begin{equation}
\begin{split}
L_k^{\text{INR}}(\boldsymbol{\theta}) =& - \frac{1}{N_k}\sum_{\boldsymbol{x}\in S_k}\log p(y=1|\boldsymbol{x};\boldsymbol{\theta}) \\
&- \sum_{j=0, j\neq k}^{K-1} \lambda_j \frac{1}{N_j}\sum_{\boldsymbol{x}\in S_j}\log (1-p(y=1|\boldsymbol{x};\boldsymbol{\theta})),
\end{split}
\end{equation}
where $p(y=1|\boldsymbol{x};\boldsymbol{\theta})$ is the predicted probability of sample $\boldsymbol{x}$ belonging to the positive class and $\lambda_j$ is the weight for class $j$ as a sub-class of the negative class. 

The first reweighting is at the level of positive \textit{vs.} negative class. If we require $\sum_j \lambda_j = 1$, using inverse-frequencies will maintain the balance between the positive and negative class, as one-\textit{vs.}-all is likely to introduce class imbalance. The second level of reweighting is within the negative class: we upweight the contribution of a hard sub-class by assigning a larger $\lambda$, making LGL-IGR focus more on the learning for that class.

\noindent\textbf{Choice of $\lambda$s} When there are a large number of classes, treating all $\lambda$s as hyperparameters and selecting the optimal values are not feasible in practice as we generally don't have the prior knowledge about which classes are hard. Instead, we adopt a strategy that assigns the weights during the training process. For each non-target class $j (j\ne k)$, let $S_j^{\text{MB}}$ be the subset of $S_j$ in the mini-batch, we use the mean predicted probability 
\[\bar{p}_j=\frac{1}{|S_j^{\text{MB}}|}\sum_{\boldsymbol{x}\in S_j^{\text{MB}}} p(y=1|\boldsymbol{x},\boldsymbol{\theta})\]
as the class-level hardness measurement. A larger $\bar{p}_j$ implies class $j$ is harder to separate from the target class $k$. We then transform those $\bar{p}_j$'s using softmax to get $\lambda_j$:
\[\lambda_j = \frac{\exp(\beta \bar{p}_j)}{\sum_{i\ne k}\exp( \beta\bar{p}_i )},\]
where $\beta \geq 0$ is the temperature that can smooth ($0\leq\beta\leq 1$) or sharpen ($\beta>1$) each non-target class's contribution \cite{chorowski2015attention}. LGL-INR adaptively shifts its learning focus to those hard classes, meanwhile keep attentive on those easy classes. Note that this strategy only introduces one extra parameter in LGL-INR.

With the competition mechanism imposed by $\sum \lambda_j=1$, LGL-INR can be viewed as a smoothed learning objective between the one-\textit{vs.}-one and one-\textit{vs.}-all approach: when $\beta=0$, $\lambda_j=1/K-1$, all non-target classes are weighted equally, which is the in-negative-class balanced LGL using inverse-class frequencies; when $\beta$ is very large, $\lambda_j$ concentrates on the hardest class (e.g., $\lambda_j\approx 1$) and LGL-INR approximately performs one-\textit{vs.}-one classification. We don't specifically fine-tune the optimal value of $\beta$ and $\beta=1$ works well in our experiments.

\section{Experiments}
We evaluate LGL-INR on several benchmark datasets for image classification. Note that in our experiments, applying LGL in multi-class classification naturally introduces data imbalance which is handled in our LGL-INR formulation. Our primary goal here is to demonstrate that LGL-INR can be used as a drop-in replacement for LGL and SML with competitive or even better performance, rather than outperform the existing best models using extra training techniques. For fair comparison, all loss functions are evaluated in the same test setting. Code is made publicly available at \textit{https://github.com/Dichoto/LGL-INR}.

\subsection{Experiment Setup}
\noindent\textbf{Dataset} We perform experiments on four MNIST-type datasets, MNIST, Fashion-MNIST (FMNIST) \cite{xiao2017/online}, Kuzushiji-MNIST (KMNIST) \cite{clanuwat2018deep} and CIFAR10. FMNIST and KMNIST are intended as drop-in replacements for MNIST which are harder than MNIST. Both datasets are gray-scale images consisting of 10 classes of clothing and Japanese character respectively. CIFAR10 consists of colored images of size $32\times32$ from 10 objects.

\begin{table}[t]
	\centering
	\begin{tabular}{c|c}  
		\hline
		Model & Architecture \\
		\hline 
		\multirow{2}{*}{CNN2C} & CV(C20K5S1)-MP(K2S2)- \\
		& CV(C50K5S1)-MP(K2S2)-800-10  \\
		\hline
		\multirow{3}{*}{CNN5C} & CV(C32K3S1)-BN-CV(C64K3S1)-BN- \\
		& CV(C128K3S1)-MP(K2S2)-CV(C256K3S1)-\\
		& BN-CV(C512K3S1)-MP(K8S1)-512-10   \\
		\hline
	\end{tabular}
	\caption{CNN architectures used for MNIST-type datasets. C-channel represents number, K-kernel size, S-stride, BN-batch normalization and MP-max pooling}
	\label{tab:cnnarch}
\end{table}

\noindent\textbf{Model setup} We test three loss functions on each dataset with different CNN architectures. For MNIST-type datasets, two CNNs with simple configurations are used. The first one (CNN2C) has two convolution layers and the other one (CNN5C) has 5 convolution layers with batch normalization \cite{ioffe2015batch}. For CIFAR10, we use MobilenetV2 \cite{howard2017mobilenets} and Resnet-18 \cite{resnet} with publicly available implementations.

\noindent\textbf{Implementation details} All models are trained with the standard stochastic gradient descent (SGD) algorithm. The training setups are as follows. For MNIST-type data, the learning rate is set to 0.01, the momentum is 0.5, batch size 64, number of epoch is 20. We don't perform any data augmentation. For CIFAR data, we train the models with 100 epochs and set batch size to 64. The initial learning rate is set to 0.1, and divide it by 10 at 50-th and 75-th epoch. The weight decay is $10^{-4}$ and the momentum in SGD is 0.9. Data augmentation includes random crop and horizontal flip. We train all models without pretraining on large-scale image data. Model performance is evaluated by the top-1 accuracy rate and we report this metric on the testing data from the standard train/test split of those datasets for fair performance evaluation. For LGL-INR, we report the results using $\beta=1$.

\subsection{Predictive Results}
Table \ref{tab:mnist-typePerf} and Table \ref{tab:cifarPerf} shows the classification accuracy using LGL, SML and LGL-INR on the MNIST-type and CIFAR10 dataset respectively. From the table, we can observe that for all three loss functions, model with larger capacity yields higher accuracy. On MNIST-type data, LGL yields overall poorer performance than SML. This is because in those datasets, some classes are very similar to each other (like shirt \textit{vs.} coat in FMNIST) and the negative class consists of 9 different sub-classes. Hence the learning focus of LGL may get distracted from the hard sub-classes due to the averaging behavior of LGL as shown in Eq (\ref{eq:keyEq2}). However, SML doesn't suffer this problem as all negative sub-classes are treated equally. On CIFAR10, LGL achieves better accuracy than SML. This is possibly due to the lack of very similar classes as in MNIST-type data. This observation demonstrates LGL's potential as a competitive alternative to SML in some classification tasks.  

\begin{table}[t]
	\centering
	\begin{tabular}{clccc}  
		\hline
		Model & Loss  &   MNIST  & FMNIST & KMNIST \\
		\hline
		\multirow{3}{*}{CNN2C} & LGL & 99.15 & 89.44 & 94.37\\
		& SML & 99.09 & \textbf{91.15} & 95.13\\
		& LGL-INR & \textbf{99.29} & \textbf{91.15} & \textbf{96.43}\\
		\hline
		\multirow{3}{*}{CNN5C} & LGL & 99.36 & 92.35 & 96.35\\
		& SML & 99.47 & 93.15 & 96.39\\
		& LGL-INR & \textbf{99.63} & \textbf{93.54} & \textbf{97.46}\\
		\hline
	\end{tabular}
	\caption{Predictive top-1 accuracy rate (\%) on the standard testing data of MNIST-type datasets.}
	\label{tab:mnist-typePerf}
\end{table}

On the other hand, LGL-INR adaptively pays more attention on the hard classes while keeps its separation from easy classes. This enables LGL-IRN to outperform LGL and SML notably. Comparing LGL-IRN with LGL, we see that the multi-modality neglect problem deteriorates LGL's ability of learning discriminative features representation, which can be relieved by the in-negative class reweighing mechanism; comparing LGL-IRN with SML, focusing on learning hard classes (not restricted to classes similar to the target class) is beneficial. Also, the adaptive weight assignment in the training process doesn't require extra effort on the weight selection, making our method widely applicable. 

\begin{table}[t]
	\centering
	\begin{tabular}{lcc}  
		\hline
		Loss & MobilenetV2 & Resnet18 \\
		\hline
		LGL  & 92.40 & 91.55\\
		SML  & 91.11 & 91.32\\
		LGL-INR & \textbf{93.34} & \textbf{93.68}\\
		\hline
	\end{tabular}
	\caption{Predictive top-1 accuracy rate (\%) on the standard testing data of CIFAR10 using different models.}
	\label{tab:cifarPerf}
\end{table}

\begin{figure*}[t]
	\centering
	\includegraphics[scale=0.52]{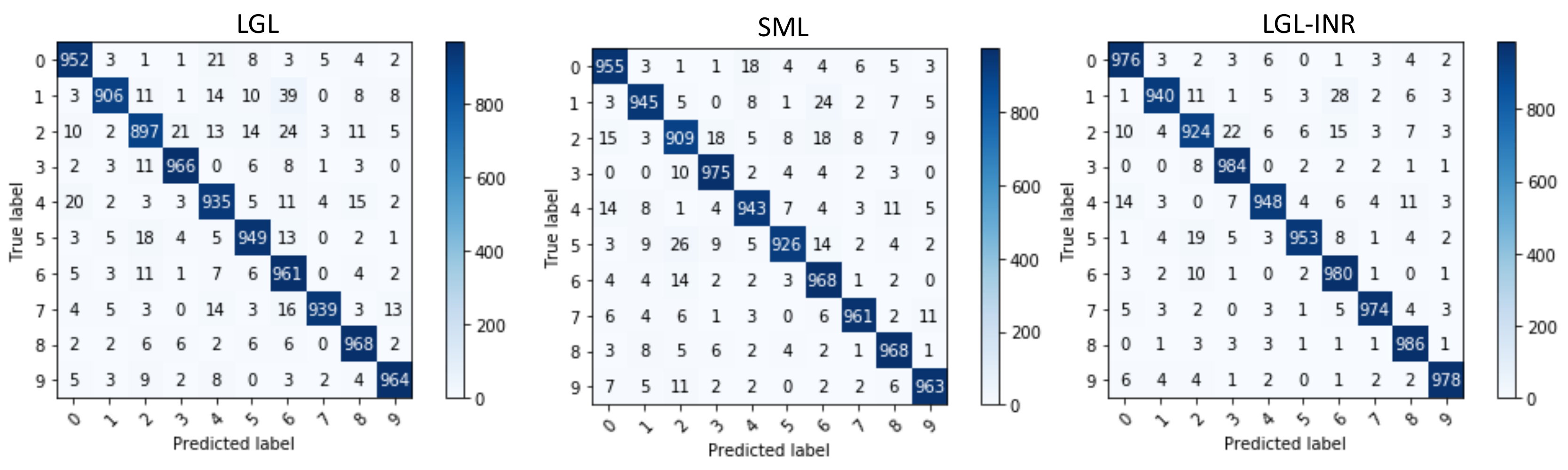}
	\caption{Confusion matrix on KMNIST testing data for LGL, SML and LGL-INR. Model: CNN2C. See Table \ref{tab:mnist-typePerf} for overall accuracy. Notably, LGL-INR outperforms LGL in all 10 classes and SML in 9 classes except  Class 1 (LGL-INR 940 \textit{vs.} SML 945), in terms of per-class accuracy.}
	\label{fig:confcm}
\end{figure*}

\begin{figure}[t]
	\centering
	\includegraphics[scale=0.5]{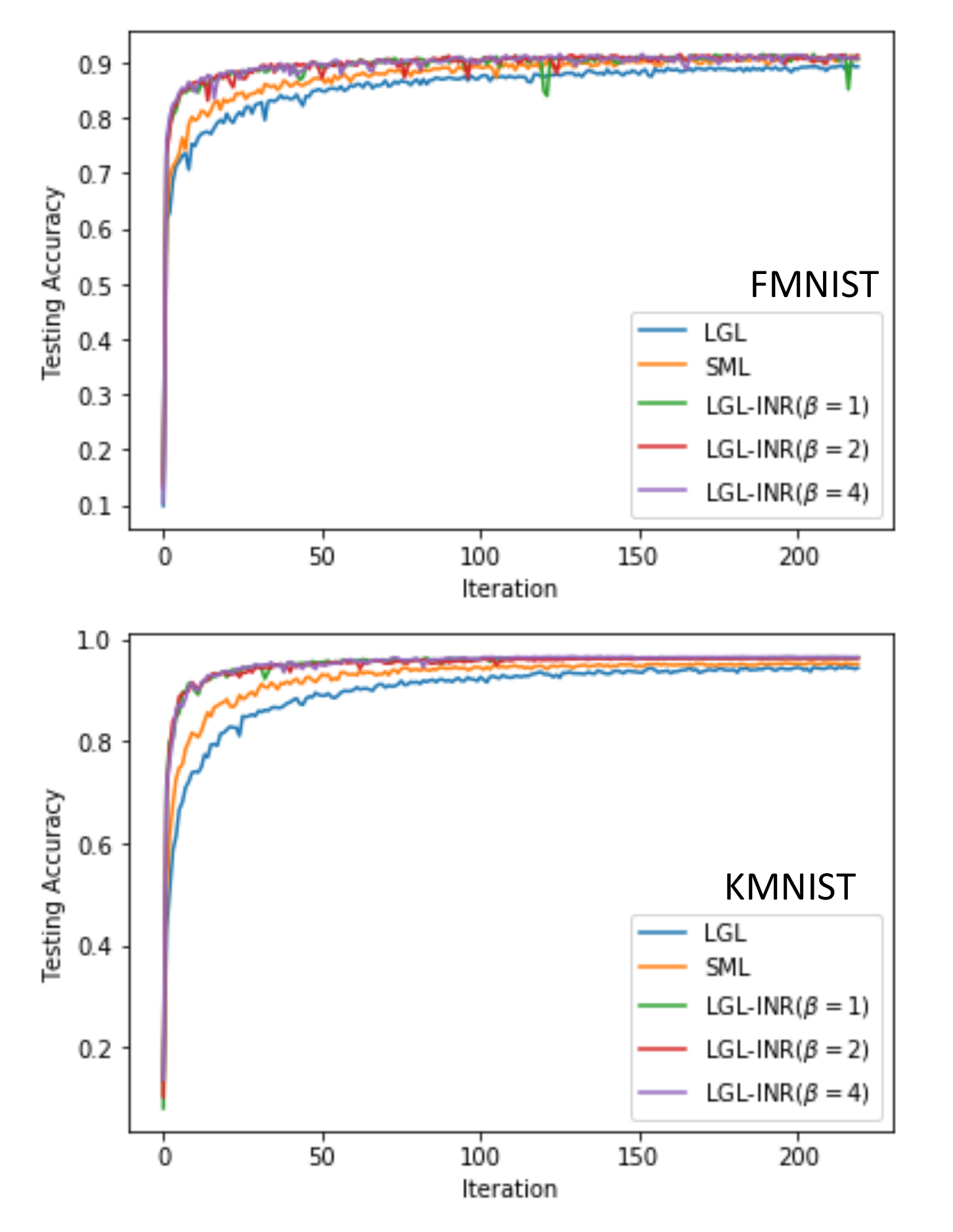}
	\caption{Testing top-1 accuracy on FMNIST and MNIST.}
	\label{fig:accCurve}
\end{figure} 

\subsection{Further Analysis}
We check the predictive behavior of LGL-INR in detail by looking at the confusion matrix on testing data. Here, we use CNN2C and KMNIST dataset as an example. Fig. \ref{fig:confcm} show the results. We observe that for LGL, Class 1 and 2 have the lowest accuracy among 10 classes. By shifting LGL's learning focus on hard classes, LGL-INR significantly improves model performance on class 1 and 2. This is within our expectation backed by the theoretical depiction of LGL's learning property. SML does not have the multi-modality neglect problem as each class is treated equally in the learning process, yet it also does not pay more attention to the hard classes. This makes LGL-INR advantageous: LGL-INR outperforms SML on 9 classes out of 10. For example, class 0 have 18 samples misclassified into class 4 whereas only 6 are misclassified in LGL-INR.

Figure \ref{fig:accCurve} displays the training accuracy curve for LGL, SML and LGL-INR on FMNIST and KMNIST. Under the same training protocol, LGL-INR achieves slightly faster convergence rate than SML and LGL with comparative (FMNIST) or better (KMNIST) performance, implying that focusing on learning hard classes may facilitate model training process.

We also check the sensitivity of the temperature parameter $\beta$ in LGL-INR weighting mechanism. Mathematically, a large or small value for $\beta$ is not desirable as the LGL-INR is reduced to an approximate one-\textit{vs.}-one  or a class-balanced learning objective. We test $\beta=1,2,4$ on KMNIST. As shown in Table \ref{tab:beta} and Fig. \ref{fig:accCurve}, model performance is not sensitive to $\beta$ in this range, making LGL-INR a competitive alternative to LGL or SML without introducing much hyper-parameter tuning.

\begin{table}[t]
	\centering
	\begin{tabular}{cccc}  
		\hline
		$\beta$  & 1 & 2 & 4\\
		\hline
		Accuracy & 96.43 & 96.29 & 96.43 \\
		\hline
	\end{tabular}
	\caption{Accuracy of different $\beta$ values on KMNIST. Model: CNN2C.}
	\label{tab:beta}
\end{table}

\section{Conclusion}
In this paper, motivated to explain the class-wise reweighting mechanism in LGL and SML, we theoretically deprived a system of probability equations that depicts the learning property of LGL and SML, as well as explains the roles of those class-wise weights in the loss function. By examining the difference in the effects of the weight mechanism on LGL and SML, we identify the multi-modality neglect problem is the major obstacle that can negatively affect LGL's performance in multi-class classification. We remedy this shortcoming of LGL with a in-negative-class reweighting mechanism. The proposed method shows its effectiveness on several benchmark image datasets. For future works, we plan to incorporate the estimation of data distribution and use the reweighting mechanism of LGL-INR at the sample level in the model training process to further improve the efficacy of the reweighting mechanism.

\section{Acknowledgement}
This work is supported by the National Science Foundation under grant no. IIS-1724227.

\bibliographystyle{aaai}
\bibliography{LRWNCRref}

\end{document}